\documentclass[conference]{IEEEtran}
\IEEEoverridecommandlockouts

\usepackage{cite}
\usepackage{amsmath,amssymb,amsfonts}
\usepackage{graphicx}
\usepackage{textcomp}
\usepackage{xcolor}
\usepackage{booktabs}
\usepackage{array}
\usepackage{tabularx}
\usepackage{colortbl}
\usepackage{url}
\usepackage{microtype}

\definecolor{PWMBlue}{HTML}{245A8D}
\definecolor{PWMTeal}{HTML}{28796F}
\definecolor{PWMGold}{HTML}{A87913}
\definecolor{PWMPurple}{HTML}{665A99}
\definecolor{PWMRed}{HTML}{9C4F54}
\definecolor{PWMInk}{HTML}{2F3437}
\definecolor{PWMPanel}{HTML}{F7FBFD}
\definecolor{PWMSoft}{HTML}{F3F6F8}
\newcolumntype{Y}{>{\raggedright\arraybackslash}X}
\newcommand{\Rone}{\textcolor{PWMTeal}{\textbf{R1}}}
\newcommand{\Rtwo}{\textcolor{PWMBlue}{\textbf{R2}}}
\newcommand{\Rthree}{\textcolor{PWMPurple}{\textbf{R3}}}
\newcommand{\Rfour}{\textcolor{PWMGold}{\textbf{R4}}}
\newcommand{\stresshead}[1]{\textcolor{PWMRed}{\textbf{#1}}}

\def\BibTeX{{\rm B\kern-.05em{\sc i\kern-.025em b}\kern-.08em
    T\kern-.1667em\lower.7ex\hbox{E}\kern-.125emX}}

\begin{document}

\title{From Event Logs to Governed Action: A BlueSky Agenda for Agentic Process Mining}

\author{
\IEEEauthorblockN{Yiyuan Yang\IEEEauthorrefmark{1},
Zheshun Wu\IEEEauthorrefmark{2},
Yong Chu\IEEEauthorrefmark{2},
Zhenghua Chen\IEEEauthorrefmark{3},
Zenglin Xu\IEEEauthorrefmark{4}, and
Qingsong Wen\IEEEauthorrefmark{5}}
\IEEEauthorblockA{\IEEEauthorrefmark{1}University of Oxford, United Kingdom \quad
\IEEEauthorrefmark{2}Peng Cheng Laboratory, Shenzhen, China}
\IEEEauthorblockA{\IEEEauthorrefmark{3}University of Glasgow, United Kingdom \quad
\IEEEauthorrefmark{4}Fudan University, China \quad
\IEEEauthorrefmark{5}Squirrel Ai Learning, United States}
\IEEEauthorblockA{ yiyuan.yang@cs.ox.ac.uk \quad wuzhsh23@gmail.com \quad
chuyong.edu@gmail.com}
\IEEEauthorblockA{ chen0832@e.ntu.edu.sg \quad zenglin@gmail.com \quad
qingsongedu@gmail.com}
\thanks{Corresponding author: Qingsong Wen
(e-mail: qingsongedu@gmail.com).}
}

\maketitle

\begin{abstract}
Process mining has long turned event logs into process knowledge: discovered models, conformance evidence, bottleneck diagnoses, and runtime predictions. Agentic AI changes the target. Process-aware agents will not only ask what happened. They will ask whether a proposed action should be taken, given the available evidence, privacy budget, organizational authority, and downstream risk. This BlueSky paper proposes \emph{event-to-action process mining}: a process-mining agenda for transforming heterogeneous operational event data into governed action. The goal is not another dashboard, a generic enterprise simulator, or a language interface over logs. We argue that the community needs four mineable artifacts: event-object representations, action evidence packages, governance contracts, and benchmarks where act, defer, ask, and refuse are all valid outputs. This agenda is timely because agentic business process management (BPM), LLM-assisted process mining, object-centric event standards, causal process monitoring, and privacy-preserving learning are maturing separately. Bringing them together defines a data-mining target inside process mining: mining logged organizational behavior for accountable action, not only retrospective insight.
\end{abstract}


\section{The BlueSky Idea}
Process mining began with a disciplined promise: if work leaves digital traces, process mining can reconstruct the real process rather than the process written in a manual. Event logs made this promise operational. They record cases, activities, timestamps, resources, objects, costs, and increasingly text or policy attributes. From them, process mining discovers models, checks conformance, predicts outcomes, and analyzes performance~\cite{vanDerAalst2016ProcessMining}. PM4Py implements discovery, conformance, and performance analysis~\cite{Berti2019PM4Py}. This makes process mining a data-mining discipline with special semantics: events are not independent samples, and a process model is not only a predictor. It is evidence about how organizations actually behave, where a mined pattern can become a managerial decision rather than only a statistical description.

\vspace{0.5em}
\noindent\begingroup\setlength{\fboxsep}{4.4pt}\fcolorbox{PWMBlue!70}{PWMPanel}{%
\begin{minipage}{0.92\columnwidth}
\textcolor{PWMBlue}{\textbf{BlueSky claim.}} The next frontier is not natural-language access to process dashboards. It is \emph{event-to-action process mining}: mining operational event data so that agents can recommend, defer, or refuse actions with causal evidence, privacy protection, and auditable authority.
\end{minipage}}
\endgroup
\vspace{0.5em}

The reason this is a BlueSky problem is that the question is changing. Classical process mining asks what happened, whether the observed behavior conforms to a model, and where improvement opportunities might lie. Agentic operations ask a sharper question: should a software or human agent reroute this case, relax this rule, call this supplier, approve this claim, escalate this patient, or wait for better evidence? A discovered Petri net, directly-follows graph, or object-centric model can describe the past. It does not by itself justify an action under uncertainty, privacy limits, organizational authority, and downstream risk.

This paper therefore defines a sharper target for process mining as a data-mining problem: moving from \emph{log-to-model} analysis to \emph{log-to-action} evidence. We are not proposing a full enterprise simulator. Event logs are partial, policy-shaped traces generated by people, software systems, incentives, and missing observations. The tractable target is to mine operational data into evidence about a specified action that exposes uncertainty and causal support, respects privacy and authority, and permits refusal when a log cannot justify intervention.

The proposal challenges three assumptions that quietly limit current work. First, prescriptive process monitoring provides a starting point for evidence about decisions affecting future process states. Second, process agents cannot be made safe by prompting alone. They need verifiable local computation, calibrated uncertainty, and explicit authority boundaries. Third, cross-organizational process mining is not merely a privacy problem. It is a setting where actions by one actor can shift cost, delay, or risk to another. The agenda is concrete for a data-mining audience: we define four mineable artifacts, \Rone{} representation, \Rtwo{} evidence, \Rthree{} governance, and \Rfour{} evaluation, that turn agentic process mining from a broad aspiration into a testable research program.

\section{Why Now: The Action Gap in Process Mining}
The timing is unusual because five research streams are now reaching the same boundary: agents can act, process mining can ground them, but the field has not yet defined what mined event evidence should authorize action.

\textbf{Agentic BPM is becoming explicit.} Recent position papers argue that generative and agentic AI may shift business process management from automation to autonomy, and from design-driven management toward data-driven management grounded in process mining~\cite{Dumas2026ABPMS}. A 2026 manifesto describes process-aware agents that act within explicit constraints, explain decisions, interact through conversation, and adapt their behavior~\cite{Calvanese2026Manifesto}. Work on digital twins of business processes similarly emphasizes real-time replicas and simulation capabilities, but also notes unresolved ambiguity in how business processes and digital twins should be integrated~\cite{Fornari2024DigitalTwins}. Formal work further emphasizes that agents need explicit goals and guardrails when autonomous decision-makers execute processes~\cite{DeGiacomo2026Formal}. These papers define the organizational need, but they leave a process-mining gap: what logged evidence is sufficient for a process agent to act?

\textbf{LLM-based process mining exposes the grounding gap.} LLM interfaces provide natural-language and code-based access to process-mining tasks~\cite{Berti2023LLMPM,Berti2024PM4PyLLM}. PMAx separates local process computations from language-based interpretation; using metadata instead of raw logs aims to mitigate metric hallucinations and limit data exposure~\cite{Antonov2026PMAx}. Domain adaptation of LLMs for process data also suggests that event sequences can become a learning substrate rather than only text prompts~\cite{Oyamada2025ProcessLLM}. The next question is not whether agents can explain a process artifact. It is whether a process artifact, a learned model, and a tool trace can jointly support a governed action.

\textbf{Object-centric logs make operational state mineable.} Traditional event logs flatten work into case traces, but real processes are networks of orders, invoices, patients, shipments, machines, documents, employees, and regulations. Object-centric logs represent many-to-many event--object relations~\cite{vanDerAalst2020OCPN}. OCEL 2.0 further supports evolving object attributes and exchange formats~\cite{Berti2024OCEL}. This matters for action: an intervention on one object can affect another object, another organization, or a resource queue. Exogenous-aware temporal prediction shows how external covariates can improve forecasts when combined with historical context~\cite{Tayal2024ExoTST}. Event-to-action mining adds process semantics, object relations, and authority constraints to the broader temporal data problem.

\textbf{Prescriptive and causal process mining reveal the evaluation problem.} Prescriptive process monitoring already studies when to intervene in running cases under uncertainty and resource constraints~\cite{shoush2022intervene}. SimBank offers controlled evaluation with simulator-known counterfactual outcomes~\cite{DeMoor2025SimBank}; ProCause studies learned generators for evaluating intervention policies~\cite{DeMoor2025ProCause}. Both address the absence of outcomes for unchosen actions in recorded logs. Causal process mining also shows that directly-follows relations can be misleading when relational event data and domain knowledge are ignored~\cite{Waibel2022CausalPM}. Event-to-action process mining builds on these methods by connecting intervention evidence to explicit privacy and authority conditions.

\textbf{Privacy and federation are no longer optional.} The most valuable processes cross organizational boundaries: hospitals and labs, insurers and providers, ports and carriers, suppliers and manufacturers, public agencies and contractors~\cite{yang2024intelligent}. Federated process mining shows that event data can be exploited across boundaries without pooling raw logs~\cite{vanDerAalst2021Federated,Khan2021CrossSilo}, while federated learning and secure aggregation support learning without centralized raw data~\cite{Kairouz2021Advances,Bonawitz2017SecureAgg}. Work on dataset similarity~\cite{Elhussein2025DatasetSimilarity} and federated fairness and privacy~\cite{Chen2025FairLDP} addresses heterogeneity and competing learning objectives. If a model guides action, privacy is not a preprocessing step. It is part of the action contract.

Together, these streams show why the process-mining gap is also a data-mining problem: mining heterogeneous event streams, learning under drift, aligning schemas, extracting causal signals from biased observational logs, recommending under uncertainty, preserving privacy, and evaluating systems when the ground truth of an unchosen action is missing. The missing output is a governed action object that binds these pieces together.

\section{From Event Logs to Governed Action}
\begin{figure}[t]
\centering
\includegraphics[width=1\columnwidth]{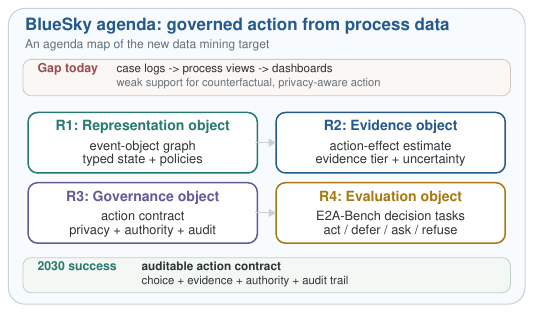}
\vspace{-20pt}
\caption{Agenda map for event-to-action process mining: mineable artifacts that make governed action computable and auditable.}
\vspace{-10pt}
\label{fig:pwm-loop}
\end{figure}

Fig.~\ref{fig:pwm-loop} is the agenda map for that shift. Today's stack often moves from case logs to process views and dashboards. The missing target is governed action. The figure names four artifacts that make this target concrete: a representation object, an evidence object, a governance object, and an evaluation object. Here, an artifact means an inspectable data structure or task specification, rather than an established process-mining term.

We call the new output a \emph{governed action recommendation}. It is not just an intervention label. It contains five parts: the proposed action, the predicted process consequences, the causal and observational evidence tier, the privacy and authority conditions under which the action is allowed, and an audit trail linking the claim to local computations or model estimates. A recommendation may be positive (act now), conditional (act only if a manager approves or a privacy budget remains), deferential (collect more evidence), or negative (refuse because the action is unsupported or unauthorized). This output type turns process mining from retrospective explanation into an action-conditioned data mining problem.

This target differs from common process mining outputs. A discovered model abstracts observed control flow, a predictive model forecasts a label, a process dashboard visualizes metrics, and a prescriptive monitor recommends an intervention for a running case. Event-to-action process mining asks for a governed evidence object that can be inspected before the action changes the organization. The agenda does not assume one model will solve all four artifacts. It defines shared interfaces for future models, tools, and benchmarks.

The target also changes in cross-organizational process mining. Many operational processes are ecosystems: a delayed shipment affects a hospital, a supplier quality issue affects insurance claims, and a fraud rule at one bank may shift behavior to another. Raw logs cannot simply be pooled, but decisions made from one partial view can still affect other actors. Federated process mining becomes one component of a larger action-governance problem rather than the endpoint.

The bet is that this target can become benchmarkable. A system should not receive credit merely for a plausible narrative or a high next-event accuracy. It should receive credit when it improves action quality under distribution shift, reports uncertainty honestly, preserves privacy, explains which evidence was used, and refuses actions that exceed its authority. The agenda is therefore not ``add an LLM to process mining'' or ``optimize one intervention policy.'' It is a call to define the four artifacts below before deployed agents normalize weaker outputs.

\subsection{Four Mineable Artifacts}
\textcolor{PWMTeal}{\textbf{R1: Representation object.}} The dependency chain starts with an explicit object that represents the operational state before any recommendation is made. A trace is too flat, and a generic embedding is too opaque. The representation object should be an event-object graph: typed objects, events, resources, queues, documents, decisions, policies, timestamps, and external conditions linked through time. It must handle concurrency, missing events, repeated work, long-tail variants, and labels whose meanings drift across organizations. Recent conformance-aware deep learning already suggests that local token accuracy can miss global process structure~\cite{DeSmedt2026DiffERO}. The data mining challenge is to learn representations that preserve process semantics: masked event reconstruction, object-state forecasting, resource contention prediction, constraint violation detection, time-to-deadline estimation, and action-effect consistency. The output of R1 is not a hidden vector alone. It is a mineable state object that can be inspected, shared selectively, and passed to evidence estimation.

\textcolor{PWMBlue}{\textbf{R2: Evidence object.}} A governed recommendation needs more than a probability. It needs an evidence object that states what is known, what is assumed, and what remains unsupported. Most event logs are observational, selective, and produced by systems optimized for execution rather than causal inference. Yet agentic operations ask causal questions: if an agent changes triage priority, supplier selection, review order, credit routing, or resource allocation, what downstream effect should it expect? The evidence object should combine temporal structure, relational objects, domain constraints, partial experiments, and realistic process generators. It should also expose evidence tiers: historical association, adjusted observational estimate, quasi-experimental support, expert assumption, generator-only conjecture, or verified intervention. SimBank and ProCause provide controlled settings for evaluating prescriptive methods with counterfactual outcomes~\cite{DeMoor2025SimBank,DeMoor2025ProCause}. R2 makes that insight part of the mining output: a recommendation carries its causal status and uncertainty, not only its score.

\textcolor{PWMPurple}{\textbf{R3: Governance object.}} If a process agent can act, the mining output must include a governance object. This object specifies which tool computed which metric, which log view was used, which causal assumption was declared, which privacy rule permitted the query, which budget was spent, and which human or organizational authority allowed the action. LLM-based process agents are useful because managers ask messy natural-language questions, but they become dangerous when they produce numbers or policies without verifiable computation. Governance therefore turns agentic process mining from conversational analytics into reproducible data mining. It also changes cross-organizational mining. An intervention learned from one partner may shift workload, delay, liability, or competitive risk to another. Secure aggregation protects individual updates within its threat model~\cite{Bonawitz2017SecureAgg}, while differential privacy formalizes limits on disclosure~\cite{Kairouz2021Advances}. Neither grants authority to act. Governed action must also report partner contribution, rare-event exposure, robustness to withheld views, and authority boundaries. The audit trail is not paperwork. It is part of the model output.

\textcolor{PWMGold}{\textbf{R4: Evaluation object.}} Research papers are easy to overclaim unless the community can test the claim. Event-to-action process mining therefore needs an evaluation object: benchmark tasks where the correct behavior may be to act, defer, ask for approval, or refuse. Learning to defer already studies transfer to an expert~\cite{Mozannar2020Defer}; here, the target also includes process evidence, privacy conditions, and action authority. Governed action should reward usefulness with restraint. Possible tasks include estimating the effect of a policy change, ranking bottleneck interventions, transferring an anomaly detector to a low-data partner, detecting a misleading natural-language request, and refusing an action because evidence, privacy budget, or authority is insufficient. Metrics should include calibration, causal validity, privacy cost, communication cost, robustness to drift, auditability, and refusal quality. A system that acts confidently on unsupported evidence should lose, even when its next-event prediction is plausible.

\section{Stress Tests and Success Criteria}
The four artifacts should be judged by failure modes, not component scores. Each stress test names an assumption that a standard process-mining improvement would leave mostly intact, and each test defines what a future process-mining system must make measurable.

\stresshead{The agenda fails if process evidence is treated as neutral telemetry.} Event logs are produced by software configurations, employee workarounds, policy incentives, missing documentation, and strategic behavior. The same event label may mean routine progress in one organization and a compliance violation in another. R1, therefore, must expose uncertain labels, contested case boundaries, and incomplete object relations rather than hide them inside an embedding.

\stresshead{The agenda fails if counterfactuals are reduced to scores.} A wrong next-event prediction degrades a forecast, and a wrong action recommendation can delay care, deny benefits, trigger compliance failures, or transfer cost to a weaker partner. Many interventions cannot be randomized at the ecosystem scale. R2 should therefore report the tier of support behind each estimate and distinguish causal evidence from persuasive but non-actionable association.

\stresshead{The agenda fails if agent safety is separated from process safety.} An agent may satisfy a formal rule while exploiting a loophole in the logging system, optimize a local service-level agreement while worsening system-level fairness, or present biased historical action as neutral efficiency. R3 must make policy constraints, privacy cost, tool provenance, approval boundaries, and risk transfer visible before action.

\stresshead{The agenda fails if deployment is evaluated as a static prediction problem.} Once agents recommend triage, supplier, exception, or resource changes, workers and partners adapt. The next event log is no longer drawn from the same distribution that trained the model. R4 must test action quality after feedback, not only prediction quality before deployment.

\stresshead{The agenda fails if refusal is counted as non-performance.} Governed process agents ask sequences of questions, invoke tools, request explanations, and refine actions. Each step can leak information. A high-quality system should refuse when event coverage is sparse, causal assumptions are unsupported, privacy budgets are exhausted, authority is missing, or an intervention would transfer unacceptable risk. Refusal is a pipeline-level property: failures in representation, evidence, or governance may all require non-action.

\noindent\textbf{What success looks like.} By 2030, success would be visible in shared artifacts: open event-to-action schemas, standard evidence tiers, privacy-aware governance protocols, and benchmarks where refusal, deferment, and approval are legitimate outputs. A mature system should transfer across low-data partners, state uncertainty and causal evidence, ground numbers in local computation, refuse unsupported actions, and expose data use, privacy cost, and approval. Evaluation should report utility gain over the current policy together with violations across complete decision episodes, including refusals.

\par\smallskip\noindent\textbf{Why this belongs in data mining.} These stress tests clarify why the agenda is broader than BPM systems or LLM interfaces. BPM contributes process semantics. LLM research contributes language interfaces, and process mining contributes event-log grounding. The data-mining contribution is to treat representation, causality, recommendation, privacy, robustness, and evaluation as one output contract over heterogeneous operational data.

The data, target, and evaluation are all unusual. Event logs are partial observations generated by humans, software, policies, and incentives. Missingness is meaningful, labels are negotiated, and distributions change when agents act on mined outputs. The output is not only a prediction, cluster, rule, or visualization, but also a governed action recommendation with uncertainty, evidence tier, privacy accounting, local verification, and authority constraints. For many actions, no single ground-truth label exists because the alternative action was never taken. Benchmarks must therefore combine logged data, realistic process generators, expert-validated counterfactuals, adversarial logs, and online or quasi-experimental evidence.

The stakes make the problem worth formalizing before deployment. Process agents will operate where errors are organizational events, not only technical failures: a bad recommendation can redistribute work, delay service, deny benefits, or hide unfairness under the language of optimization. Governed action turns responsibility into a data mining test: models must know when prediction is insufficient.

\noindent\textbf{A first community challenge.} The agenda can begin with a shared benchmark. A first public data-mining challenge, \emph{E2A-Bench}, could release realistic process ecosystems with partial object-centric logs, local tool APIs, privacy budgets, intervention catalogs, and hidden ground-truth mechanisms. Participants would answer action-conditioned questions about policy changes, resource reassignment, withheld partner views, and forbidden action classes. Table~\ref{tab:e2abench} turns the four artifacts in Fig.~\ref{fig:pwm-loop} into an agenda matrix where act, defer, ask, and refuse are all scored outcomes. A fixed, versioned core supports reproducibility, while renewed hidden scenarios and sequential feedback test adaptation. Results should report benchmark versions separately to distinguish changing difficulty from progress.

\begin{table}[t]
\caption{E2A-Bench scoring matrix for governed process action.}
\label{tab:e2abench}
\centering
\footnotesize
\setlength{\tabcolsep}{2.0pt}
\renewcommand{\arraystretch}{1.10}
\begin{tabularx}{\columnwidth}{@{}>{\raggedright\arraybackslash}p{0.25\columnwidth}>{\raggedright\arraybackslash}p{0.38\columnwidth}Y@{}}
\toprule
\rowcolor{PWMSoft}
\textbf{Artifact} & \textbf{Mining target} & \textbf{Success signal}\\
\midrule
\rowcolor{PWMTeal!6}
\Rone{} Representation & Typed event-object state with policies and temporal validity. &
Object-aware answers remain stable under schema drift.\\
\rowcolor{PWMBlue!6}
\Rtwo{} Evidence & Action-effect estimates with uncertainty and explicit evidence tiers. &
Causal support is separated from association.\\
\rowcolor{PWMPurple!6}
\Rthree{} Governance & Action contract with tool trace, privacy cost, authority, and audit trail. &
Action is verifiable, approved, and audit-ready.\\
\rowcolor{PWMGold!8}
\Rfour{} Evaluation & Decision tasks for act, defer, ask, and refuse under drift. &
Calibrated restraint is rewarded when action is unsafe.\\
\bottomrule
\end{tabularx}
\end{table}
\section{Conclusion}

Process mining was built on the premise that event logs can reveal how work actually unfolds. This paper extends that premise to the agentic setting: when mined evidence is used to trigger or block organizational action, the output must carry evidence, constraints, and accountability together. Event-to-action process mining therefore requires four connected artifacts: representations, evidence packages, governance contracts, and evaluation tasks that turn logs into recommendations that an organization can verify, limit, defer, or refuse. Success is not only automation. It includes deferral or refusal when logs are incomplete, evidence is weak, or authority is absent. The central message is simple: the next frontier is not a larger dashboard or an unconstrained process agent, but a data-mining foundation for accountable process action.

\clearpage
\IEEEtriggeratref{14}
\bibliographystyle{IEEEtran}
\bibliography{6_ref}

\end{document}